\documentclass{article}

\usepackage{microtype}
\usepackage{graphicx}
\usepackage{subfigure}
\usepackage{booktabs} 
\usepackage{hyperref}

\newcommand{\bs}{\bar{S}}

\usepackage[accepted]{icml2020}

\usepackage{mystyle}

\usepackage{color}

\icmltitlerunning{Go Wide, Then Narrow: Efficient Training of Deep Thin Networks}

\begin{document}

\global\long\def\E{\mathbb{E}}%
\global\long\def\x{\textbf{x}}%
\global\long\def\y{\text{y}}%
\global\long\def\z{\text{\textbf{z}}}%
\global\long\def\M{\text{\textbf{M}}}%

\global\long\def\th{\boldsymbol{\theta}}%
\global\long\def\thtnil{\th_{t}^{N,i,\ell}}%
\global\long\def\thtil{\bar{\th}_{t}^{i,\ell}}%

\global\long\def\ztnl{\z_{t}^{N,\ell}}%
\global\long\def\ztl{\z_{t}^{\infty,\ell}}%

\global\long\def\rtl{\rho_{t}^{\infty,\ell}}%
\global\long\def\rtnl{\rho_{t}^{N,\ell}}%
\global\long\def\rtn{\rho_{t}^{N,[1,L]}}%
\global\long\def\rt{\rho_{t}^{\infty,[1,L]}}%
\global\long\def\brtnl{\bar{\rho}_{t}^{N,\ell}}%

\global\long\def\lip{\text{Lip}}%

\twocolumn[
\icmltitle{Go Wide, Then Narrow: Efficient Training of Deep Thin Networks}



\icmlsetsymbol{equal}{*}

\begin{icmlauthorlist}
\icmlauthor{Denny Zhou}{go}
\icmlauthor{Mao Ye}{ut}
\icmlauthor{Chen Chen*}{go}
\icmlauthor{Tianjian Meng*}{go}
\icmlauthor{Mingxing Tan*}{go}
\icmlauthor{Xiaodan Song}{go}
\icmlauthor{Quoc Le}{go}
\icmlauthor{Qiang Liu}{ut}
\icmlauthor{Dale Schuurmans}{go}
\end{icmlauthorlist}

\icmlaffiliation{go}{Google Brain, USA}
\icmlaffiliation{ut}{Department of Computer Science, The University of Texas at Austin, USA}

\icmlcorrespondingauthor{Denny Zhou}{dennyzhou@google.com}

\icmlkeywords{Deep Learning, Model Compression}

\vskip 0.3in
]



\printAffiliationsAndNotice{\icmlEqualContribution} 

\begin{abstract}
%
For deploying a deep learning model into production, it needs to be both accurate and compact to meet the latency and memory constraints. This usually results in a network that is deep (to ensure performance) and yet  thin (to improve computational efficiency). In this paper, we propose an efficient method to train a deep thin network with a theoretic guarantee. Our method is motivated by model compression. It consists of three stages. First, we sufficiently widen the deep thin network and train it until convergence. Then, we use this well-trained deep wide network to warm up (or initialize) the original deep thin network. This is achieved by layerwise imitation, that is, forcing the thin network to mimic the intermediate outputs of the wide network from layer to layer.  Finally, we further fine tune this already well-initialized deep thin network. The theoretical guarantee is established by using the neural mean field analysis. It demonstrates the advantage of our layerwise imitation approach over backpropagation.  We also conduct large-scale empirical experiments to validate the proposed method. By training with our method, ResNet50 can outperform  ResNet101, and $\text{BERT}_\text{BASE}$ can be comparable with $\text{BERT}_\text{LARGE}$, when  ResNet101 and $\text{BERT}_\text{LARGE}$ are trained under the standard training procedures as in the literature. 
\end{abstract}

\section{Introduction}
\label{sec:int}

In many machine learning applications, in particular,  language modeling and image classification, it is becoming common to dramatically increase the model size to achieve significant performance improvement \citep[e.g.,][]{he2016deep, wu2016google,  devlin2018bert, brock2018large, raffel2019exploring, brown2020language}. To enlarge a model, we can make it either much deeper or wider. 
A big deep learning model may involve millions or even billions of parameters, and be trained over a big computation cluster containing hundreds or thousands of computational nodes. 

Despite their impressive performance, however, it is almost impossible to directly deploy these big deep learning models into production because of the low latency and memory constraints. To remedy this issue, there has been an increasing interest in developing compact versions of good performing big models to meet the practical constraints while without much drop of accuracy \citep[e.g.,][]{iandola2016squeezenet, sandler2018mobilenetv2, howard2019searching, efficientnet2019, wortsman2019discovering, zhiqing2020mobile}.  

%

%
In general, compact modeling results in networks which are both deep and thin.  This is because a compact model must be sufficiently deep in order to extract hierarchical high-level representations that are impossible for shallow models to accomplish \citep[e.g.,][]{lee2009convolutional, le2012building,  allen2020backward},
but each layer of the model does not have to be very wide since many neurons are redundant and can be pruned without hurting the performance \citep[e.g.,][]{han2015learning, li2017pruning, frankle2019lottery, liu2019rethinking, ye2020good}.


Nevertheless, training deep thin networks are much more difficult than training deep wide networks. 
The loss surface of a deep thin network tends to be highly irregular and nonconvex \citep[e.g.,][]{li2018visualizing}. Moreover,  during the backpropagation through a deep thin network,  gradients may vanish or explode  \citep[e.g.,][]{bengio1994learning, pascanu2013difficulty}.  On the other hand, it has been observed that increasing the width of, or ``overparameterizing'' the network 
makes it much easier to train since its loss surface becomes smoother and nearly convex \citep[e.g.,][]{li2018visualizing, du2018gradient, allen2019convergence}. 

\begin{figure*}[tb]
\centering
\includegraphics[width=0.56\textwidth]{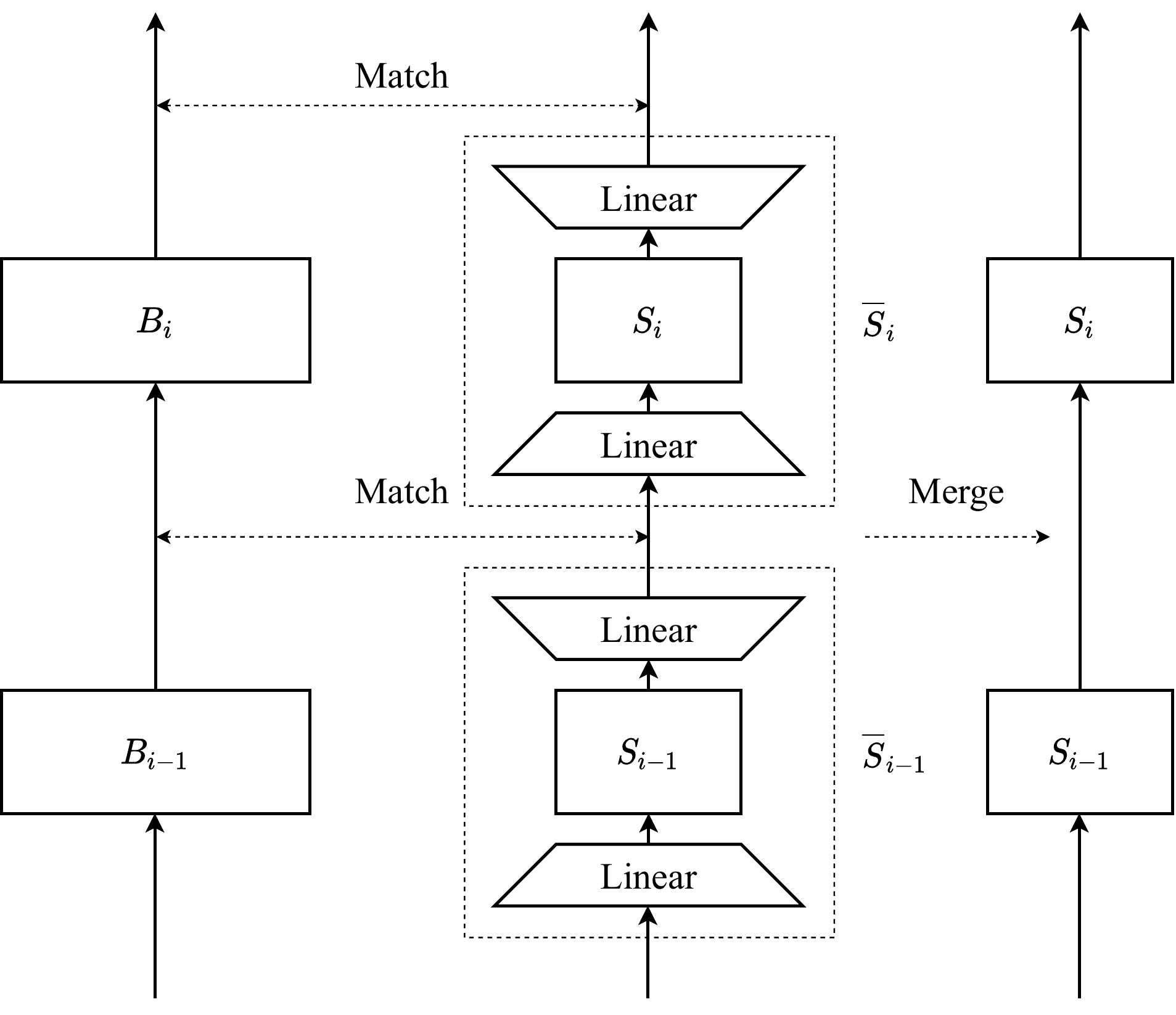} 
\caption{Illustrating ``go wide, then narrow". Left panel: wide network $B$ with building blocks $B_i$ obtained by widening $S_i$ in the thin network $S$. Middle panel:  network $\bs$  obtained by inserting appropriately sized linear transformation pairs between any two adjacent building blocks $S_{i-1}$ and $S_i$ in $S$.  Right panel: merge all adjacent linear transformations in $\bs$ to restore its architecture to $S$.} 
\label{fig:match}
\end{figure*}

In this paper, we propose a generic algorithm to train deep thin networks with a theoretical guarantee. Our method is motivated by model compression.  It consists of three stages (Figure \ref{fig:match} and Algorithm \ref{alg:pt}). In the first stage, we significantly widen the deep thin network to obtain a deep wide network, for example, twice wider, and then train it until convergence. In the second stage, we use this well trained deep wide network to warm up (or initialize) the original deep thin network.  In the last stage, we further fine tune this already well-initialized deep thin network.  Since a deep thin network is  highly nonconvex,  a good initialization is almost all we need to obtain a good training result.

The key component of our method lies in  its second stage. In this stage, the deep thin network is gradually warmed up by layer-to-layer imitating the intermediate outputs of the well trained deep wide network obtained in the first stage in a bottom-to-top fashion. 
This is analogous to curriculum learning, where a teacher breaks down an advanced learning topic to a sequence of small learning tasks, and then students learn these small tasks one by one under the teacher's stepwise guidance. 

The readers may have noticed a technical issue to conduct layerwise imitation learning: the thin network and its wide version differs in the output dimension in every layer. This causes the difficulty to measure how well the thin network mimics the wide one. Obviously, such a problem does not exist in knowledge distillation since the final output dimension stays the same for either network.  For example, the final outputs could be one-dimensional labels. 
To fix this dimension mismatch issue in our layerwise imitation learning, we insert a pair of linear  transformations between any two adjacent layers in the thin network (see Figure~\ref{fig:match}): one is used to increase the dimension of its layer output to match the dimension of the corresponding layer output of the wide network, and the other to reduce the dimension back to its original size. 
Thus, the dimension expanded output from the thin network can be compared with the output from the wide network from layer to layer using elementwise metrics like the mean squared loss and Kullback-Leibler (KL) divergence. 
Since a sequence of linear layers is mathematically equivalent to a single linear layer,  at the end of our algorithm, we can merge all adjacent linear layers in the thin network,  including the linear transformations inside the original network modules. Thus, the thin network architecture is eventually restored to its original design.  
%

We develop theoretical analysis for our method using the mean field analysis of neural networks \cite{song2018mean,  araujo2019mean, nguyen2020rigorous}. 
We show that, compared with direct gradient descent training of a deep thin network, 
our method allows for much simpler and tighter error bounds (see Proposition \ref{prop:disc} vs. Theorem \ref{thm:main}). 
The  intuition underlying the theoretical analysis  is that our layerwise imitation scheme avoids backpropagation through the deep network and consequently prevents an explosive growth of the error bound on the network depth.  
Similar theoretical results do not hold for 
the commonly used knowledge distillation and its variants \citep[e.g.,][]{ba2014deep, hinton2015distilling}, because they only modify the training target to include a  distillation loss  but still depend on backpropagation through the entire deep thin  network. 

In additional to theoretic analysis, we also conduct large scale empirical experiments to validate our approach.
we train the ResNet \cite{he2016deep}  and BERT \cite{devlin2018bert} models using our method and the baseline methods which include training  deep thin networks from scratch via backpropagation and training with knowledge distillation.
Experimental results show that our method significantly outperforms the baseline methods. In particular, by training with our method, ResNet50 can outperform  ResNet101,  and $\text{BERT}_\text{BASE}$ can be comparable with $\text{BERT}_\text{LARGE}, $ when  ResNet101 and $\text{BERT}_\text{LARGE}$ are trained under the standard training procedures as in the literature.

We organize this paper as follows. In Section \ref{sec:alg}, we present our algorithm for training deep thin networks. In Section \ref{sec:the}, we develop theoretic results around our method using the mean field analysis for neural networks. The proof details are provided in Appendix. The work related to our algorithm and theoretic analysis are discussed in Section \ref{sec:the}. In Section \ref{sec:exp}, we present the experimental results of training the ResNet and BERT models using different methods. Finally, we conclude this paper with discussions in Section \ref{sec:con}.

\section{Algorithm}
\label{sec:alg}

Let us denote by $S = \{S_1, S_2, \ldots, S_n\}$ a deep  thin network that we want to train,  where $S_i$ denotes the building block at the $i$-th layer of $S.$ 
For an input $x$, the output of network $S$ is given as $S(x) = (S_n \circ  S_{n-1} \circ \cdots \circ S_1)(x) = S_n(S_{n-1}( \cdots S_2(S_1(x))\cdots)).$ 

In a feed forward network, a building block can be a hidden layer or a group of hidden layers. However, in many other neural architectures, the structure of a building block can be much more complex. Here are two typical examples. In a model for image classification, a building block usually contains convolution, pooling, and batch normalization \cite{he2016deep}; in a model for language modeling, a building block may include multi-head attention, feed forward network, and layer normalization \cite{vaswani2017attention}.

\textbf{Stage 1: Wide learning}. In this stage, we first construct a deep wide network $B = \{B_1, \dots, B_n\}, $ where building block $B_i$ is obtained by significantly \emph{widening} building block $S_i$ in the deep thin network $S.  $ We then train this deep wide network $B$ until convergence. In general, the wider the network,  the easier to train it. How to make a  network wider may not be that straightforward as it looks like. It is actually fairly case dependent. For a feed forward network, we can widen it by introducing more neurons in its hidden layers;  for a convolution network, we can widen it by introducing more filters;  for a transformer like model, we can widen it from multiple dimensions, including increasing its hidden dimension, using more self-attention heads, or adding more hidden neurons in its feed forward network module.

\textbf{Stage 2: Narrow learning}.  In this stage, we first construct a new network $\bs$ by inserting a pair of \emph{appropriately  sized} linear transformations  $\{M_{i, 1},  M_{i, 2}\}$ between  two adjacent building blocks $S_{i}$ and $S_{i+1}$ in the thin network $S$ (see the middle column of Figure \ref{fig:match} for an illustration of $\bs$). The first linear transformation $M_{i, 1}$ increases the output dimension of $S_i$ to match the output dimension of $B_i$ in the wide network $B, $ and the second linear transformation $M_{i, 2}$ reduces the dimension to its original size. 
Formally, the new network $\bs$ can be  written as 
$\bs = \{S_1, M_{1, 1}, M_{1, 2}, S_2, \dots, S_{n-1},  M_{n-1, 1}, M_{n-1, 2},  S_n\}. $ 

We group the modules in $\bs$ as follows:  $\bs_1 = \{S_1, M_{1, 1}\},$ $\bs_i = \{M_{i-1, 2}, S_i, M_{i, 1}\}$ for $i = 2, \dots,  n-1, $ and  $\bs_n = \{M_{n-1, 2}, S_n\}. $  Thus,  $\bs = \{\bs_1, \bs_2, \dots, \bs_n\}.  $ 
Next, for $i = 1, \dots, n-1, $  we sequentially train a set of subnetworks $\bs^{(i)} = \{\bs_1, \dots, \bs_i\}$ 
by minimizing the output discrepancy between $\bs^{(i)}$ and  subnetwork $B^{(i)} = \{B_1, \dots, B_i\}$ in the wide network $B.$ The instances in the training data are used as the inputs. Note that the weights of $B^{(i)}$ are fixed since the entire network $B$ has been trained in the first stage. In addition, during this sequential training, the trained $\bs^{(i)}$ is naturally served as initialization when proceeding to training $\bs^{(i+1)}. $ There are many ways to measure the output discrepancy between $\bs^{(i)}$ and $B^{(i)}$. Typical choices include the mean squared error and KL divergence. 

To achieve a better performance, when each training subnetwork $\bs^{(i)}$,  we may restart multiple times. In each restart, the most recently added building block $S_i$ is randomly reinitialized. Finally, we choose the trained network which best mimics $B^{(i)}$ before proceeding to training $\bs^{(i+1)}.$

\begin{algorithm}[tb]
   \caption{go WIde, then Narrow (WIN)}
   \label{alg:pt}
\begin{algorithmic}
   \STATE {\bfseries Input:} thin network $S=\{S_1, \dots, S_n\}$; training data 
   \STATE
   \STATE \textbf{ Stage 1: Wide learning}. Construct a wide network $B=\{B_1,  \cdots, B_n\}, $ where each $B_i$ is obtained by widening  $S_i$ in network $S$, and train $B$ until convergence.
   \STATE
   \STATE \textbf{Stage 2: Narrow learning}. Construct another network $\bs=\{\bs_1, \dots, \bs_n\}, $ where each $\bs_i$ is obtained from  wrapping up $S_i$  by two appropriately sized linear transformations  to match the output dimension of $B_i.$   
   \STATE{}
   \FOR{$i=1$ {\bfseries to} $n - 1$}
   \STATE Train subnetwork $\bs^{(i)} = \{\bs_1, \dots, \bs_i\}$ by minimizing the output discrepancy between $\bs^{(i)}$ and  subnetwork $B^{(i)} =\{B_1, \dots, B_i\}$ from the wide network $B$.
   \ENDFOR
   \STATE
   \STATE \textbf{Stage 3: Fine-tuning and merging}. Use the training labels to fine tune network $\bs^{(n)} = \bs$, and then merge all adjacent linear layers in $\bs$ to restore its architecture to $S.$
\end{algorithmic}
\end{algorithm}

\textbf{Stage 3: Fine-tuning and merging}. This is the final stage of our method. After layerwise imitation in the second stage,  the network $\bs^{(n)} = \bs$ has been well initialized. We can further fine tune this network using the training labels. Afterwards, we merge all adjacent linear transformations in $\bs$, including the native linear layers residing in its building blocks,  Consequently, $\bs$ is restored to the architecture of the original deep thin network $S$ (see the illustration in Figure \ref{fig:match}). Until then our algorithm is done. Optionally, one may restart fine-tuning several times,  and then choose the model which has the minimum training error. Such a greedy choice usually works well since a thin model is supposed to have a strong regularization effect on its own.  

We summarize our method in Algorithm \ref{alg:pt}. Here are several additional remarks. First, our method does not have to be more expensive than the normal knowledge distillation method since the imitation training in our method is quite light in each layer. In addition, we often can use an existing trained big model as the wide network in our method. Consequently, the wide learning stage of our method can be skipped. Moreover, the layers in our method do not have to exactly align to the layers of the trained neural models. For example, the wide network may be a 24-layer $\text{BERT}_\text{LARGE}$ model, and the narrow network a 12-layer $\text{BERT}_\text{BASE}$ model. Then, during the imitation training, each layer in $\text{BERT}_\text{BASE}$ is mapped to two adjacent layers in $\text{BERT}_\text{LARGE}$.

\section{Theoretical Analysis}
\label{sec:the}
In this section, we present a theoretical comparison between
a layerwise imitation based scheme and the standard gradient descent training.  
The basic intuition underlying our analysis is that  
layerwise imitation breaks the learning of a deep thin network into a sequence of shallow subnetworks training, and  hence avoids backpropagation through the entire deep thin  network from top to bottom. This makes our method more suitable for training very deep networks, and also enables simpler theoretical analysis.  
Our theoretic results show that 
layerwise imitation yields a much tighter error bound compared with gradient descent. 

\subsection{Assumptions and Theoretical Results}
\label{sec:technical}
Our analysis is built on the theory of mean field analysis of neural network \citep[e.g.,][]{song2018mean, araujo2019mean, nguyen2020rigorous}. We start with the formulation of deep mean field network formulated by \citet{araujo2019mean}. 

For notational conventions, let 
$S^{m}=\left\{ S_{1}^m,S_{2}^m,\cdots,S_{n}^m\right\}$ 
and 
$B^{M}=\left\{ B_{1}^M,B_{2}^M,\cdots,B_{n}^M\right\}$ 
be the thin and wide networks of interest, where 
we add the superscribes $m$ and $M$ to denote the number of neurons in each layer of the thin and wide networks, respectively. 
We assume the $i$-th layer of $S^{m}$ and $B^{M}$ are 
\begin{align*} 
S_{i}^m(\z)=\frac{1}{m}\!\sum_{j=1}^{m}\!\sigma\!\!\left(\z,\th_{i,j}^S\right), && 
B_{i}^M(\z)=\frac{1}{M}\!\sum_{j=1}^{M}\!\sigma\!\!\left(\z,\th_{i,j}^B\right),
\end{align*}
where 
$\th_{i,j}^S$ and $\th_{i,j}^{B}$ are the weights of the thin and wide models, respectively. 
 Here we also define 
\[
\sigma(\z,\th)=\th_{1}^{\top}\sigma_{+}(\z^{\top}\th_{0}),\ \ \ \th=[\th_{0},\th_{1}],
\]
where $\sigma_{+}(\cdot)$ is some commonly used nonlinear element-wise mapping such as sigmoid. 

In order to match the dimensions of the thin and wide models, 
we 
assume the input and output of both $S_{i}^m(\z)$ and $B_{i}^M(\z)$ of all the layers (except the output)  have the same dimension $d$, so that $\z \in \RR^d$ and $\th_{i,j}^S, \th_{i,j}^B\in\R^{(d+1)\times d}$ for all the neurons and layers.
In practice, this can be ensured by inserting linear transform pairs as we have described in our practical algorithm (so that the $S_i^m$ corresponds to the $\bar S_i$ in Figure\ref{fig:match}). 
In addition, for the sake of simplicity, the dimension of the final output is assumed to be one.  

Giving a dataset $\mathcal D = \{\x_i, y\}_i$, 
we consider the regression problem of 
minimizing the mean squared error: 
\begin{align}\label{equ:lossmse}
\mathcal{L}(F)=\E_{(\x,y)\sim \mathcal D}\left[\left(F(\x)-y\right)^{2}\right],
\end{align}
via gradient descent with step size $\eta$ and a proper random initialization. We define the output discrepancy between two models $S$ and $B$ to be 
\[
\D[S,B]=\sqrt{\mathbb{E}_{\x\sim \mathcal D}\left [\left(S(\x)-B(\x)\right)^{2} \right]}.
\]

\begin{ass} \label{asm:init}
Denote by $B_{\gd}^m$ and $S_{\gd}^M$ the result of running stochastic gradient descent on dataset $\mathcal D$ with a constant step size $\eta>0$, for a fixed $T$ steps. 

For both models,  
we initialize  the parameters $\{\th_{i,j}^S\}_{j\in [m]}$ and $\{\th_{i,j}^B\}_{j\in [M]}$ in the $i$-th layer by drawing i.i.d. samples from a distribution $\rho_i$. 
We suppose $\rho_i$ is absolute continuous and has bounded domain for $i\in[n]$.
\end{ass}

\paragraph{Bounds of Deep Thin Networks Trained from Scratch}

Analyzing deep neural networks trained with gradient descent remains a ground challenge in theoretical deep learning. 
The few existing bounds \citep{araujo2019mean, nguyen2020rigorous} depend rather poorly on the depth of networks, due to the difficulty of controlling the errors propagated through the layers during gradient descent. 
Here we leverage the mean field analysis from \citet{araujo2019mean} to give an estimation of the discrepancy between the thin and wide models $S_{\gd}^m$ and $B_{\gd}^M$.

\begin{ass} \label{asm:bound}
Suppose the data and labels in $\mathcal D$ are bounded, i.e. $\left\Vert \x\right\Vert \le c$ and $\left|y\right|\le c$ for some $c<\infty$. And suppose the activation function $\sigma_+$ and its first and second derivatives are bounded.
\end{ass}
%
The boundedness assumptions are typical and (almost) always required in the theoretical literature on deep learning \citep[e.g.,][]{song2018mean,araujo2019mean}.  Relaxing the boundedness assumptions could be possible but it brings more technical issues while without bringing additional insights. In practice, given that the size of observed data is finite, we can always assume that the data and weights are properly truncated and thus bounded. Actually, for image data,  they are already bounded, and the trainable weights are usually initialized from a truncated distribution.


\begin{pro} \textbf{(Discrepancy between the wide and thin networks trained by stochastic gradient descent)}
Under assumption \ref{asm:bound} and \ref{asm:init}, we have 
\begin{align*} 
&\D[S^m_{\gd},B^M_{\gd}] \\
&=\mathcal{O}_p\left(n\exp c_1(\exp(c_2 n))
{\small \left  (\frac{1}{\sqrt{m}} +\frac{1}{\sqrt{M}}+\sqrt{\eta}\right )}\right),
\end{align*} 

where $c_1,c_2>0$ are some positive constant, $\mathcal O_p(\cdot)$ denotes the big O notation in probability, and the randomness is w.r.t. the random initialization of gradient descent, and the random mini-batches of stochastic gradient descent. 
\label{prop:disc}
\end{pro}

The proof of this bound is 
based on the proof of Theorem 5.5 in \citet{araujo2019mean}; see Appendix for details. Because the $m$ is small and $n$ is large for deep thin networks,
the bound above is dominated by $n\exp(c_1\exp(c_2n))m^{-1/2}$, which decreases with the width $m$, but grows double exponentially with the depth $n$. 
The poor dependency on $n$ is both due to the critical gradient vanish/exploding problem when backpropagating through deep networks and the mathematical challenge for analyzing deep networks under the mean field framework.  

\paragraph{Breaking the Curse of Depth with Layerwise Imitation} 
Now we show how our layerwise imitation algorithm can help training deep thin networks.

\begin{ass} \label{asm:init_2}
Denote by $S^m_{\mathrm{WIN}}$ the result of mimicking $B_{\gd}^M$ following Algorithm~\ref{alg:pt}.  
When training $S_{\mathrm{WIN}}^m$, 
we assume the parameters of  $S^m_{\mathrm{WIN}}$ in each layer are initialized by 
randomly sampling $m$ neurons from the
the corresponding layer of 
the wide network $B_{\gd}^M$. 
Define  $B^M_{\gd,[i:n]}=B_n^{M}\circ \cdots B_{i}^M$. 
\end{ass}

\begin{thm} \textbf{(Main Result)}
Assume all the layers of $B^M_{\gd}$ are Lipschitz maps and all its parameters are bounded by some constant.  Under Assumption 
\ref{asm:init}, 
\ref{asm:bound},
\ref{asm:init_2},
we have 
\[
\D[S_{\mathrm{WIN}}^m, B^M_{\gd}]
=
\mathcal{O}_p\left(\frac{\ell_B n}{\sqrt{m}}\right),
\]
where $\ell_B = \max_{i\in[n]}\norm{B^M_{\gd,[i+1:n]}}_{\mathrm{Lip}}$ and $\mathcal O_p(\cdot)$ denotes the big O notation in probability, and the randomness is w.r.t. the random initialization of gradient descent, and the random mini-batches of stochastic gradient descent. 
\label{thm:main}
\end{thm}

The bound above depends on linearly on $n$ and the maximum Lipschitz constant $\ell_B$. 
Because 
it is expected that the wide network  
$B^{M}_{\gd}$ is easy to train and can closely  approximate the  underlying true map, the Lipschitz constant $\ell_B$ can be mostly depended on the true map in practice (rather than how deep $B_{\gd}^M$ is) and  does not explode rapidly with $n$ like the bound on $\D[S_{\mathrm{GD}}^m, B^M_{\gd}]$. 
An important future work is to develop rigours bounds for $\ell_B$. 

\section{Related Work}
\label{sec:rel}
Our method is deeply inspired by MobileBERT \cite{zhiqing2020mobile}, which is a highly compact BERT variant designed for mobile applications with extreme memory and latency constraints. In its architecture design, the original BERT building block is replaced with a thin bottleneck structure. To train it,  MobileBERT is first initialized by imitating  the outputs of a well trained large BERT from layer to layer and then fine tuned. The main difference between MobileBERT and the proposed method is that in MobileBERT linear transformations are introduced with the bottleneck structures so they are part of the model and cannot be cancelled out by merging as in our method. In addition, the method here is generic and can be applied to any model training. 

 FitNets \cite{romero2014fitnets} also aim at training deep thin networks. In this work, a deep student network is first partially initialized by matching the output from its some chosen layer (guided layer) to the output from another chosen layer (hint layer) of a shallow teacher network. The  chosen guided and hint layers do not have to be at the same depth since the teacher network is chosen to be much shallower than the student network. After the partial initialization, the whole student network is trained via knowledge distillation. The matching is implemented by minimizing a parameterized  mean squared loss in which a parameterized regressor is applied to project the student's output such that the size of its output can match the size of the teacher's output.  The major difference between FitNets and our method is that the introduced regressor in FitNets is not part of the student network architecture. It is discarded after training. 

This kind of teacher-student paradigm can be traced back to knowledge distillation and its variants \cite{ba2014deep,hinton2015distilling}. The basic idea in knowledge distillation is to use both true labels and the outputs from a cumbersome model to train a small model. In the literature, the cumbersome model is usually referred to as teacher, and the small model student. The loss based on the teacher's outputs, that is,  the so-called distillation loss,  is linearly combined with the true labels based training loss as the final objective to train the student model. In the variants of knowledge distillation, the intermediate outputs from the teacher model are further used to construct the distillation loss which is parameterized as in FitNets. Unlike knowledge distillation, our method uses a teacher model to initialize a student model rather than constructing a new training objective. After the initialization, the student model is trained as usual. Based on such a special initialization manner, we are able to establish a theoretic guarantee for our approach.  

Our theory is built upon the mean field analysis for neural networks, which is firstly proposed by \citet{song2018mean} to study two-layer neural networks and then generalized to deep networks by \citet{araujo2019mean}. The general idea of mean field analysis is to think of the network as an interacting particle system,  and then study how the behavior of the network converges to its limiting case (as the number of neurons increases). It is  shown by \citet{araujo2019mean} that as the depth of a network increases, the stochasticity of the system increases at a double exponential scale with respect to its depth. This characterizes the problem of gradient explosion or vanish. On the other hand, they also establish the results which suggest that increasing the width of the network helps the propagation of gradient, as it reduces the stochasticity of the system. In our method, we  first train a wide network that helps the propagation of gradients,  and then force the thin network to mimic the wide network from layer to layer. Consequently, the negative influence of depth decreases from double exponential to linear.

\section{Experiments}
\label{sec:exp}
We conduct empirical evaluations by training state of the arts neural network models for image classification and natural language modeling. Our baselines include vanilla training methods for these models as shown in the literature as well as knowledge distillation. In addition, in what follows, following the convention in the literature and for the sake of convenience, we refer to the wide model in our method as teacher, and the thin model as student. 

\subsection{Image Classification}
We train the widely used  ResNet models \cite{he2016deep} on the ImageNet dataset \cite{imagenet15} using our apporach and baseline methods. 

\subsubsection{Setup}


\paragraph{Models. } 
ResNet is build on a list of bottleneck layers \cite{he2016deep}. Each bottleneck layer consists of three modules: a projection 1x1 convolution to reduce the channel size to 1/4 of the input channels, a regular 3x3 convolution, and a final expansion 1x1 convolution to recover the channel size. The wide teacher model used in our method is constructed by  increasing the channel size of the 3x3 convolution as in \cite{zagoruyko2016wide}, and the remaining two 1x1 convolutions simply keep the increased channel size without projection or expansion.  

The models that we evaluate include ResNet50, ResNet101 and their reduced versions: ResNet50-1/2, ResNet50-1/4, ResNet101-1/2 and ResNet101-1/4. For each model's reduced version, we apply the same reducing factor to all layers in that model. For example, ResNet50-1/2 means that the channel size of every layer in this model is half the channel size of the corresponding layer in ResNet50. The complexity numbers including FLOPs and parameter sizes for different models are collected in Table \ref{tab:imagenet_complexity} for reference.

\paragraph{Vanilla training setting.} We follow the training settings in \cite{he2016deep}. Each ResNet variant is trained with 90 epochs using SGD with momentum 0.9, batch norm decay 0.9, weight decay 1e-4, and batch size 256. The learning rate is  linearly increased from 0 to 0.1 in the first 5 epochs, and then reduced by 10x at epoch 30, 60 and 80.

\paragraph{WIN setting. }  We naturally split ResNet into four big chunks or building blocks with respect to the resolution change,  that is, with separations at conv2\_x, con3\_x, conv4\_x, and con5\_x. In the first stage of our method, the teacher network is constructed as 4x larger (in terms of the channel size of the 3x3 convolutions) than the corresponding student network, and trained with the vanilla setting. In the second stage,  for training each building block in the student network, we run 10 epochs by minimizing the mean squared error between the output of the teacher and student network. The optimizer is SGD with momentum 0.9. The learning rate decayed from 0.1 to 0 under the cosine decay rule. After that, we fine tune the student network for 50 epochs by minimizing the Kullback-Leibler divergence from the teacher logits to student logits, with the learning rate decayed from 0.01 to 0 under the cosine decay rule. Note that the total number of training epochs here is 90, which is the same as in the vanilla training.  We do not apply weight decay in the last two stages since the compact architecture of a thin network has already implied a strong regularization. 

\begin{table}  
    \caption{ 
    Model complexity comparison between the teacher and student models.
   }
    \vskip 0.1in        
    \centering
    \resizebox{0.95\linewidth}{!}{
        \begin{tabular}{l|cc|cc}   
           \multicolumn{1}{c}{}  &   \multicolumn{2}{c}{Teacher} &  
 \multicolumn{2}{c}{Student} \\
         &  FLOPs & Params & FLOPs & Params  \\
        \midrule
        ResNet50     & 11B  & 68M & 4.1B & 26M \\
        ResNet50-1/2 & 2.9B & 18M  & 1.1B & 6.9M\\
        ResNet50-1/4 & 0.75B & 4.7M  & 0.29B & 2.0M \\
        \midrule
        \midrule
        ResNet101   & 23B & 127M & 7.9B & 45M \\
        ResNet101-1/2 & 5.8B & 32M & 2.0B & 12M \\
        ResNet101-1/4 & 1.5B & 8.3M & 0.53B & 3.2M \\
        \end{tabular}                
    }          
    \label{tab:imagenet_complexity}       
\end{table} 

\subsubsection{Results} The evaluation results are collected in Table \ref{tab:imagenet_performence}. The numbers listed in the table cells are the top-1 accuracy on the ImageNet dataset from the models trained by different methods: our method, the vanilla training,  and knowledge distillation.  The results show that our method significantly outperforms the baseline methods. Moreover, we would like to point out that ResNet50 trained by our method achieves an accuracy of $78.4\%$ which is even higher than the accuracy  of $77.5\%$ from  ResNet101 trained by the vanilla approach.

We conduct an ablation study to demonstrate the effect of the teacher model size.  The results are shown in Table \ref{tab:imagenet_teacher}. For ResNet50, the 2x teacher performs almost equally well as the 4x teacher. The same observation holds for their students. However, for the thinner models ResNet50-1/2 and ResNet50-1/4,  the models trained by the 2x teacher are worse than the models trained by the 4x teacher. We do not try an even larger teacher such as the 6x one because of the computational cost.  


\begin{table}  
    \caption{ 
     ImageNet top-1 accuracy $(\%)$ by the models trained by the vanilla setting, knowledge distillation (KD), and our method. 
   }
    \vskip 0.1in        
    \centering
    \resizebox{0.70\columnwidth}{!}{
        \begin{tabular}{l|ccc}   
        Model & Vanilla & KD & WIN \\
        \midrule
        ResNet50        &   76.2 & 76.8 &  \textbf{78.4}  \\
        ResNet50-1/2    &   72.2 & 72.9 &  \textbf{74.6}  \\
        ResNet50-1/4    &   64.2 & 65.1 &  \textbf{66.4}  \\
        \midrule
        \midrule
        ResNet101        &   77.5 & 78.0 &  \textbf{79.1}  \\
        ResNet101-1/2    &   74.6 & 75.5 &  \textbf{76.8}  \\
        ResNet101-1/4    &   68.1 & 69.1 &  \textbf{69.7}  \\
        \end{tabular}
    }
    \label{tab:imagenet_performence}
\end{table}

\begin{table}  
    \caption{ 
    ImageNet top-1 accuracy $(\%)$ by different size teachers and their students trained with our method. 
   }
    \vskip 0.1in        
    \centering
    \resizebox{0.95\columnwidth}{!}{
        \begin{tabular}{l|cc|cc}   
         & \multicolumn{2}{c}{2x } & \multicolumn{2}{c}{4x} \\
         & Teacher & Student & Teacher & Student \\
        \midrule
        ResNet50     & 78.4 & 78.4 & 78.6 & {78.2} \\
        ResNet50-1/2 & 76.0& 74.6 & 77.6 & {75.0} \\
        ResNet50-1/4 & 70.5 & 66.4 & 74.4 & {67.5} \\
        \end{tabular}
    }
    \label{tab:imagenet_teacher}
\end{table}


\begin{figure}[tb]
\centering
\includegraphics[width=0.48\textwidth]{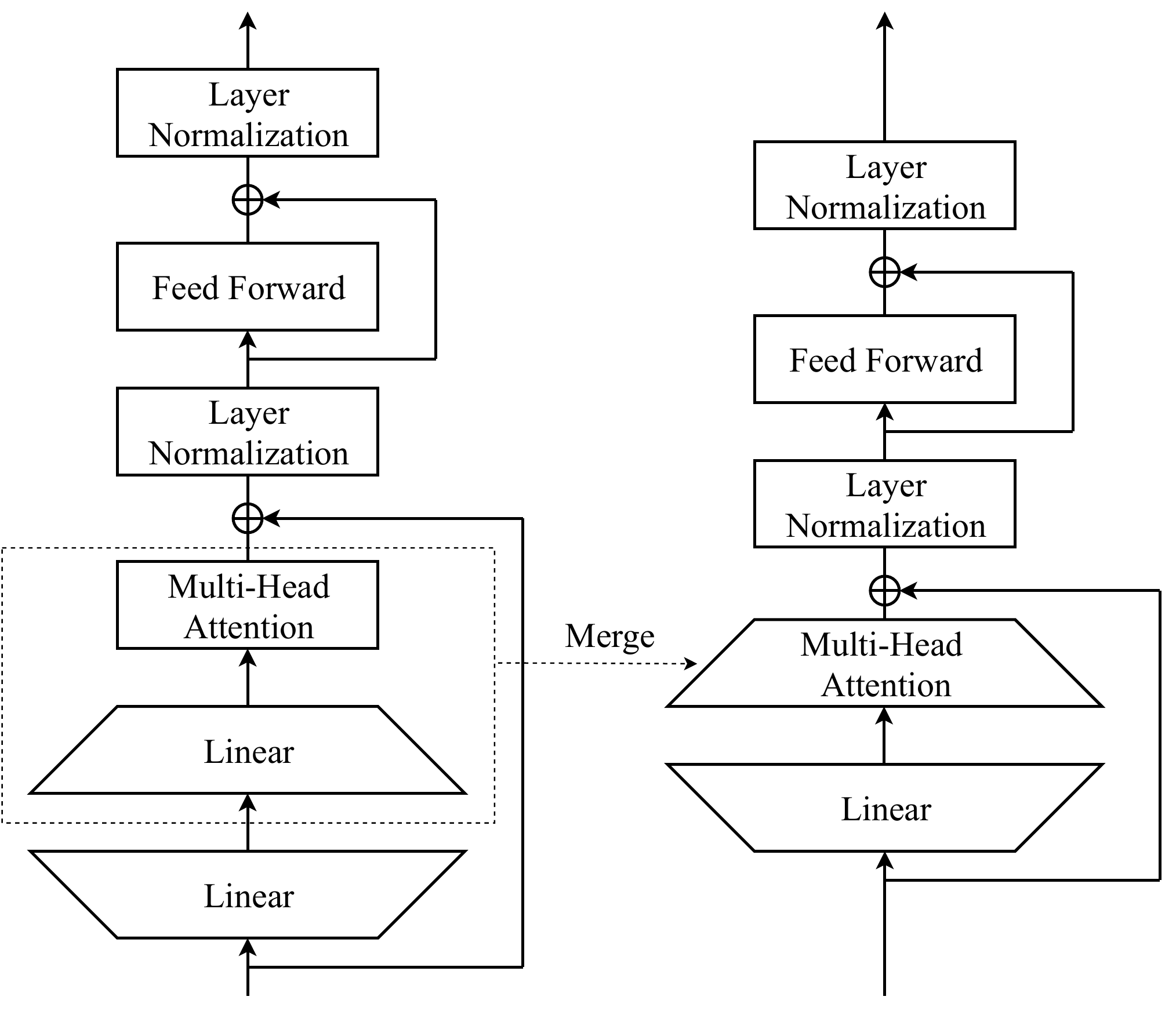}
\caption{Adding linear transformation pairs into a thin BERT model. Left panel: a pair of linear transformation are inserted between any two adjacent transformer blocks. Right Panel: the linear transformation right next to the multi-head attention module (see the dashed box in the left panel) is merged before the training in the second stage, i.e., the stage of narrow learning, while the remaining linear transformation will be merged after fine-tuning when the whole training procedure is done. Thus, finally, the trained model has the exact same network architecture and number of parameters as the original thin model. }
\label{fig:bert}
\end{figure}

\subsection{Language Modeling}
In this task, we train BERT \cite{devlin2018bert}, the state-of-the-art pre-training language model, using our method as well as the baseline methods as in the image classification tasks.  Following \citet{devlin2018bert},
we firstly pre-train the BERT model using BooksCorpus \cite{zhu2015aligning} and the Wikipedia corpus. Then we fine-tune this pre-trained model and evaluate on 
the Stanford Question Answering Dataset (SQuAD) 1.1 and 2.0 \cite{rajpurkar2016squad}.

\subsubsection{Setup}

\begin{table*}
    \caption{
    The results on the SQuAD dev datasets from the BERT models trained by our method, vanilla training and knowledge distillation (KD). \dag marks our runs with the official code.
    }
    \label{tab:squad}
    \vskip 0.1in    
    \centering
        \begin{tabular}{l|cc|cc}   
           &   \multicolumn{2}{c}{SQuAD 1.1} &  \multicolumn{2}{c}{SQuAD 2.0} \\
           Model & Exact Match & F1 & Exact Match & F1  \\
           \midrule
           BERT$_{\text{BASE}}$ \cite{devlin2018bert} & 80.8 & 88.5 & \text{~~74.2\dag} & \text{~~77.1\dag} \\
           BERT$_{\text{LARGE}}$ \cite{devlin2018bert} & 84.1 & 90.9 & 78.7 & 81.9 \\
           \midrule
           \midrule
           Teacher & 85.5 & 91.9 & 80.3 & 83.2 \\
           \midrule
           BERT$_{\text{BASE}}$ (Vanilla) & 83.6 & 90.5 & 77.9 & 80.4 \\
           BERT$_{\text{BASE}}$ (KD) & 84.2 & 90.8 & 78.9 & 81.4 \\
           BERT$_{\text{BASE}}$ (WIN)  & \textbf{85.5} & \textbf{91.8} & \textbf{79.6} & \textbf{82.5} \\
        \end{tabular}
\end{table*}

\paragraph{Models.}  The model that we are going to train here is BERT$_{\text{BASE}}$.  It takes token embeddings as its inputs and contains 12 transformer blocks \cite{vaswani2017attention}.
Each transformer block consists of one multi-head self-attention module and one feed forward network module, which are followed by layer normalization and connected by skip connections respectively.
On top of the transformer blocks, there is a classifier layer to make task-specific predictions.

The teacher model for our method is constructed by simply doubling the hidden size of every transformer block and also the width of every feed forward module in BERT$_{\text{BASE}}$. We keep the size of the teacher model's embedding the same as BERT$_{\text{BASE}}$'s and add a linear transformation right after teacher model's embedding to match its hidden size. Thus, the student model described below and the wider teacher model can share the same token embeddings as their inputs.

The way to construct the student model is illustrated in Figure \ref{fig:bert}. Specifically, taking the canonical BERT$_{\text{BASE}}$ model, we insert a pair of linear transformations between any two adjacent transformer blocks. We also put one extra linear layer over the last transformer block of BERT$_{\text{BASE}}$.
The output size of the lower linear transformation is designed to be the same as the output size of teacher model's transformer block, i.e., the teacher model's hidden size.
To more efficiently train this student model, before the training, we merge the upper linear transformation into the fully-connected layers inside the multi-head attention module. After training, we can further merge the remaining lower linear transformation into the multi-head attention module. Similarly, we can also merge the extra linear transformation over the last transformer block into the final classifier layer.
Hence, the final student model has the exact same network architecture and number of parameters as BERT$_{\text{BASE}}$.

\paragraph{Vanilla training setting.} 
There are two training phrases for the BERT models: pre-training and fine-tuning. 
In the pre-training phrase, we train the model on the masked language modeling (MLM) and next sentence prediction (NSP) tasks
using BookCorpus and Wikipedia corpus for 1 million steps with batch size of 512 and sequence length of 512.
We use the Adam optimizer with the learning rate of 1e-4, $\beta_1$ = 0.9,
$\beta_2$ = 0.999, weight decay of 0.01. The learning rate is linearly warmed up in the first 10,000 steps, and then linearly decayed. 
After pre-training, we enter the fine-tuning phrase. In this phrase, we fine tune all the parameters using the labeled data for a specific downstream task. 


\paragraph{WIN setting. } In the first stage of our method, we train the 2x wider teacher model using the vanilla method. In the second stage, we first
copy the teacher model's token embeddings to the student model,  and then progressively warm up the student's transformer blocks from layer to layer. In each step of this stage, we minimize the mean squared error between the output of the linear transformation after the student's transformer block, and the output of the teacher's corresponding transformer block. We train the first transformer block for 10k steps, the second for 20k steps until the 12th for 120K steps. Note that BERT$_{\text{BASE}}$ has 12 transformer block layers in total.   Now we enter the fine-tuning stage. We follow the same vanilla training setting to pre-train this warmed-up model on MLM and NSP tasks. Finally, we fine tune the model for downstream tasks (no knowledge distillation is employed here).

\subsubsection{Results}

We evaluate the models using the SQuAD 1.1 and 2.0 datasets. Results are shown in Table~\ref{tab:squad}. Note that  BERT$_{\text{BASE}}$ trained using our vanilla setting here outperforms BERT$_{\text{BASE}}$ \cite{devlin2018bert} by a large margin. The reason for the improvement is that we pre-train the model with sequence length of 512 for all steps, while \citet{devlin2018bert} pre-train the model with sequence length of 128 for 90\% of the steps and sequence of 512 for the rest 10\% steps. The better training result establishes a stronger baseline.
BERT$_{\text{BASE}}$ trained by our method further beats this stronger baseline by 1.9 exact match score and 1.3 F1 score on SQuAD 1.1, and 1.7 exact match score and 2.1 F1 score on SQuAD 2.0. Actually, BERT$_{\text{BASE}}$ trained by our method is comparable with BERT$_{\text{LARGE}}$ by vanilla training. 

\begin{table}
    \caption{
    The results on the SQuAD 1.1 dev dataset from the  BERT$_{\text{BASE}}$-1/2 models trained by our method and baselines.
    }
    \label{tab:squad-half}
    \vskip 0.1in    
    \centering
        \begin{tabular}{l|cc}   
           Model & Exact Match & F1 \\
           \midrule
           BERT$_{\text{BASE}}$-1/2 (Vanilla) & 78.9 & 86.3  \\
           BERT$_{\text{BASE}}$-1/2 (KD) & 80.1 & 87.4  \\
           BERT$_{\text{BASE}}$-1/2 (WIN) & 81.4 & 88.6  \\
        \end{tabular}
\end{table}

We also run experiments with a thinner student model called BERT$_{\text{BASE}}$-1/2 which halves the hidden size and width of the feed-forward network of BERT$_{\text{BASE}}$ in every layer. As shown in Table~\ref{tab:squad-half}, BERT$_{\text{BASE}}$-1/2 trained by our method significantly surpasses the same model trained by the vanilla method and knowledge distillation.

\begin{table}
    \caption{
    The results on the SQuAD 1.1 dev dataset, comparing whether to merge linear transformations after the fine-tuning phrase (MAF) or pre-training phrase (MAP) in our method.
    }
    \label{tab:squad-merge}
    \vskip 0.1in    
    \centering
        \begin{tabular}{l|cc}   
           Model & Exact Match & F1 \\
           \midrule
           BERT$_{\text{BASE}}$ (MAF) & 85.5 & 91.8 \\
           BERT$_{\text{BASE}}$ (MAP) & 85.1 & 91.5 \\
           \midrule
           \midrule
           BERT$_{\text{BASE}}$-1/2 (MAF) & 81.4 & 88.6 \\
           BERT$_{\text{BASE}}$-1/2 (MAP) & 81.4 & 88.5 \\
        \end{tabular}
\end{table}

In addition, we conduct an ablation study to demonstrate the effect of the timing for merging linear transformations. In our approach, we suggest to merge all adjacent linear layers after the fine-tuning stage when the whole training procedure is done. One may notice that,  alternatively,  we can merge the linear layers right after the narrow learning stage. So this will be before the fine-tuning stage. By using either of these two merging methods, the network structure and model size are the same. We compare these two merging methods and present results in Table~\ref{tab:squad-merge}. From the comparison,  merging after fine-tuning seems to have slightly better results. The improvement are minor but consistent. 

\section{Conclusion}
\label{sec:con}
We proposed a general method for efficiently training deep thin networks. Our method can be simply described as ``go wide, then narrow". A theoretic guarantee is developed for our method by using mean field analysis for neural networks. Empirical results on training image classification and language processing models demonstrate the advantage of our method over these two baseline methods: training deep thin networks from scratch using backpropagation as in the literature, and training with the state of the art knowledge distillation method. Our method is complimentary to existing model compression techniques including quantization and knowledge distillation. One can combine our method with these techniques to obtain an even better compact model.  For the future work, we are interested at searching for a different initialization or optimization method which is not teacher based while still enjoying a similar theoretic guarantee. If we can make it, we will be able to save the cost of training a large teacher model.




\bibliography{main}
\bibliographystyle{icml2020}
\clearpage
\onecolumn

\appendix

\section{Proof of Proposition~\ref{prop:disc}} 

This result directly follows Theorem 5.5 in \citet{araujo2019mean}. Let $B^\infty_{\gd}$ denote the infinitely wide network trained by gradient descent in the limit of  $M\to\infty$.  
By the results in Theorem 5.5 of \citet{araujo2019mean}, 
we have 
$$
\D[S_{\gd}^m, ~~ B^\infty_{\gd}] = \mathcal O_p\left ( n\exp(c_1\exp(c_2n)) \left (\frac{1}{\sqrt{m}} + \sqrt{\eta}  \right ) \right ),
$$
where we explicitly give the dependency of constant $C_{5.5}$ in \citet{araujo2019mean} on the depth $n$,  because $C_{5.5} = O(\exp(c_1\times C_{B.16}))$, where $C_{B.16}=\mathcal{O}(\exp(c_2 n))$ and $c_1$ is some positive constant. See Lemma 12.2 in \citet{araujo2019mean} for details. 

Similarly, 
$$
\D[S_{\gd}^m, ~~ B^\infty_{\gd}] = \mathcal O_p\left ( n\exp(c_1\exp(c_2n)) \left (\frac{1}{\sqrt{M}} + \sqrt{\eta}  \right) \right ). 
$$
Combining this, we have 
\begin{align*} 
\D[B_{\gd}^M, ~~ B^M_{\gd}] 
& \leq \D[S_{\gd}^m, ~~ B^\infty_{\gd}]  + 
\D[B_{\gd}^M, ~~ B^\infty_{\gd}]  \\
& = \mathcal O_p\left ( n\exp(c_1\exp(c_2n)) \left (\frac{1}{\sqrt{m}} + \frac{1}{\sqrt{M}} + \sqrt{\eta}  \right) \right ). 
\end{align*}

\section{Proof of Theorem~\ref{thm:main}}

\paragraph{Assumption~\ref{asm:init_2}} 
\emph{Denote by $S^m_{\mathrm{WIN}}$ the result of mimicking $B_{\gd}^M$ following Algorithm~\ref{alg:pt}.  
When training $S_{\mathrm{WIN}}^m$, 
we assume the parameters of  $S^m_{\mathrm{WIN}}$ in each layer are initialized by 
randomly sampling  $m$ neurons from the
the corresponding layer of 
the wide network $B_{\gd}^M$. 
Define  $B^M_{\gd,[i:n]}=B_n^{M}\circ \cdots B_{i}^M$. }

\paragraph{Theorem~\ref{thm:main}} 
\emph{Assume all the layers of $B^M_{\gd}$ are Lipschitz maps and all its parameters are bounded by some constant.  Under the  assumptions 
\ref{asm:init}, 
\ref{asm:bound},
\ref{asm:init_2},
we have 
\[
\D[S_{\mathrm{WIN}}^m, B^M_{\gd}]
=
\mathcal{O}_p\left(\frac{\ell_{B} n}{\sqrt{m}}\right),
\]
where $\ell_{B} = \max_{i\in[n]}\norm{B^M_{\gd,[i+1:n]}}_{\mathrm{Lip}}$ and $\mathcal O_p(\cdot)$ denotes the big O notation in probability, and the randomness is w.r.t. the random initialization of gradient descent, and the random mini-batches of stochastic gradient descent.}

\begin{proof} 
To simply the notation, we denote $B_{\text{GD}}^{M}$ by $B^{M}$ and  $S_{\text{WIN}}^{m}$ by $S^{m}$ in the proof. We have 
\begin{align*}
B^{M}(\x) & =(B_{n}^{M}\circ B_{n-1}^{M}\circ...\circ B_{1}^{M})(\x)\\
S^{m}(\x) & =\left(S_{n}^{m}\circ S_{n-1}^{m}\circ...\circ S_{1}^{m}\right)(\x).
\end{align*}
We define 
\[
B_{[k_1:k_2]}^{M}(\z)=(B_{k_2}^{M}\circ B_{k_2-1}^{M}\circ...\circ B_{k_1}^{M})(\z),
\]
where $\z$ is the input of $B_{[k_1:k_2]}^{M}$. 
Define 
\begin{align*}
F_{0}(\x) & =\left(B_{n}^{M}\circ...\circ B_{3}^{M}\circ B_{2}^{M}\circ B_{1}^{M}\right)(\x) \\
F_{1}(\x) & =\left(B_{n}^{M}\circ...\circ B_{3}^{M}\circ B_{2}^{M}\circ S_{1}^{m}\right)(\x)\\
F_{2}(\x) & =\left(B_{n}^{M}\circ...\circ B_{3}^{M}\circ S_{2}^{m}\circ S_{1}^{m}\right)(\x)\\
 & \cdots\\
F_{n}(\x) & =\left(S_{n}^{m}\circ...\circ S_{3}^{m}\circ S_{2}^{m}\circ S_{1}^{m}\right)(\x), 
\end{align*}
following which we have $F_{0}=B^{M}$ and $F_{n}=S^{m}$, and hence 
\[
\D[S^m, B^M] = \D[F_{n},F_{0}]\le\sum_{k=1}^{n}\D[F_{k},F_{k-1}].
\]
Define $\ell_{i-1}:=\norm{B_{[i:n]}^M}_{\lip}$ for $i\in [n]$ and $\ell_{n}=1$.  
Note that 
\begin{align*} 
\D[F_{1},F_{0}]
&  = \sqrt{\E_{\x \sim \mathcal D}\left [\left(B_{[2:n]}^M\circ B_{1}^{M}(\x)- B_{[2:n]}^M\circ S_{1}^{m}(\x)\right)^{2}\right]} \\
& \leq 
\ell_1 
\sqrt{\E_{\x \sim \mathcal D}\left [ \left (B_{1}^{M}(\x)- S_{1}^{m}(\x)\right)^{2}\right]} \\
\end{align*}
By the assumption that we initialize $S_{1}^{m}(\x)$  by randomly
sampling neurons from $B_{1}^{M}(\x)$, we have, with high probability,
\[
\sqrt{\E_{x\sim\mathcal{D}}\left [ \left(B_{1}^{M}(\x)-S_{1}^{m}(\x)\right)^{2}\right]}\le
\frac{c}{\sqrt{m}},
\]
where $c$ is constant depending on the bounds of the parameters of $B^M$. 
Therefore, 
$$
\D[F_1, F_0] = \mathcal O_p\left (\frac{\ell_1}{\sqrt{m}}\right).
$$
Similarly, we have 
\[
\D[F_{k},F_{k-1}]=
\mathcal{O}\left(\frac{\ell_{k}}{\sqrt{m}}\right), ~~~~~
\forall k = 2,\ldots, n. 
\]
Combine all the results, we have 
\[
\D[B^{M},S^{m}]=\mathcal{O}\left(\frac{n \max_{k\in[n]}\ell_k}{\sqrt{m}}\right).
\]
\end{proof} 
\paragraph{Remark} Since 
the wide network $B_{\gd}^M$ is observed to be easy to train, it is expected that it can closely approximate the underlying true function and 
behaves nicely,  hence yielding a small $\ell_B$.
An important future direction is to develop rigorous theoretical bounds  for controlling $\ell_{B}$.

\end{document}